%% file: root.tex
\documentclass[letterpaper, 10 pt, conference]{ieeeconf}   %
\UseRawInputEncoding

\IEEEoverridecommandlockouts                              %
\usepackage[backend=bibtex,bibstyle=ieee,citestyle=numeric-comp,maxbibnames=3,maxcitenames=3,doi=false,url=false,eprint=false]{biblatex}
\usepackage{hyperref}

\usepackage{graphicx} %
\usepackage{svg}
\usepackage{amsmath} %
\usepackage[font=footnotesize]{subcaption}
\usepackage{siunitx}

\usepackage{lmodern}

\title{\LARGE \bf
One Stack to Rule them All: \\
To Drive Automated Vehicles, and Reach for the  4\textsuperscript{th} Level
}

\author{
Sven Ochs$^{1}$,
Jens Doll$^{1}$,
Daniel Grimm$^{1}$,
Tobias Fleck$^{1}$,
Marc Heinrich$^{1}$,
Stefan Orf$^{1}$, \\
Albert Schotschneider$^{1}$,
Helen Gremmelmaier$^{1}$,
Rupert Polley$^{1}$,
Svetlana Pavlitska$^{1}$, 
Maximilian Zipfl$^{1}$,\\
Helen Schneider$^{2}$,
Ferdinand M{\"u}tsch$^{2}$,
Daniel Bogdoll$^{1}$,
Florian Kuhnt$^{1}$,\\
Philip Sch\"orner$^{1}$,
Marc Ren\'{e} Zofka$^{1}$
and J. Marius Z\"ollner$^{1,2}$%
\thanks{$^{1}$ Department of Technical Cognitive Systems, FZI Research Center for Information Technology, Germany.
	{\tt\small \{surname\}@fzi.de}}%
\thanks{$^{2}$ Karlsruhe Institute of Technology (KIT), Germany.}
 }

\begin{document}

\maketitle
\thispagestyle{empty}
\pagestyle{empty}

\begin{abstract}
    \input{sec/0_abstract}
\end{abstract}

\input{sec/1_intro}
\input{sec/3_architecture}
\input{sec/4_vehicles}
\input{sec/5_demo}

\input{sec/6_conclusion}

\printbibliography
\end{document}

%% file: sec/0_abstract.tex
Most automated driving functions are designed for a specific task or vehicle. Most often, the underlying architecture is fixed to specific algorithms to increase performance. Therefore, it is not possible to deploy new modules and algorithms easily. In this paper, we present our automated driving stack which combines both scalability and adaptability. Due to the modular design, our stack allows for a fast integration and testing of novel and state-of-the-art research approaches.
Furthermore, it is flexible to be used for our different testing vehicles, including modified EasyMile EZ10 shuttles and different passenger cars. These vehicles differ in multiple ways, e.g. sensor setups, control systems, maximum speed, or steering angle limitations. Finally, our stack is deployed in real world environments, including passenger transport in urban areas.
Our stack includes all components needed for operating an autonomous vehicle, including localization, perception, planning, controller, and additional safety modules. Our stack is developed, tested, and evaluated in real world traffic in multiple test sites, including the Test Area Autonomous Driving Baden-W\"{u}rttemberg.

%% file: sec/1_intro.tex
\section{INTRODUCTION}
\label{sec:introduction}

Automated driving introduces high requirements to the underlying system architecture. Given the non-linear and often unpredictable nature of research, it is critical that such an architecture is both adaptable and modular, allowing it to accommodate a wide range of potential requirements and developments. Such flexibility is essential in facilitating the development and deployment of state-of-the-art algorithms.
In addition to the aforementioned demands, automated driving functions must also strive for reproducible results that are widely generalizable and can be effectively applied to a variety of vehicles and sensor setups.

The most prominent \textit{open source} systems are \textit{Autoware}~\cite{autoware_2022_github, kato_open_2015, Kato:2018} and \textit{Apollo}~\cite{apollo_2022_github}. \textit{Autoware} was the first framework that included components for perception, prediction, planning, and control. It is based on ROS~2 and was demonstrated in a wide variety of scenarios~\cite{zang2022winning}. \textit{Autoware} is subdivided into Core modules, managed by a foundation, and Universe modules, which come on top and are community driven. Multiple projects are based on \textit{Autoware}~\cite{ASLAN_2023_Web, mehrXCARExperimentalVehicle2022}. \textit{Apollo} has a similar scope as \textit{Autoware}. It is based on its own middleware, called CyberRT. Some functionalities are only provided by closed-source cloud services. \textit{Apollo} comes with over 25 pre-trained neural networks (NN) for traffic light and lane perception, obstacle detection, and trajectory prediction. The underlying training data are not necessarily provided, which makes it hard to reuse the models for other vehicle setups~\cite{kesslerBridgingGapOpen2019a, kesslerMixedIntegerMotionPlanning2023}. Compared to \textit{Autoware}, \textit{Apollo} appears to be more robust but suffers from a smaller ecosystem~\cite{rajuPerformanceOpenAutonomous2019a}. 
Further open source projects, such as Pylot~\cite{gog_pylot_2021_icra} or \mbox{AVStack}~\cite{hallyburton_avstack_2022}, did not gain much attraction so far.

\begin{figure}[t!]
	\centering
	\begin{center}
		\includegraphics[width=\columnwidth]{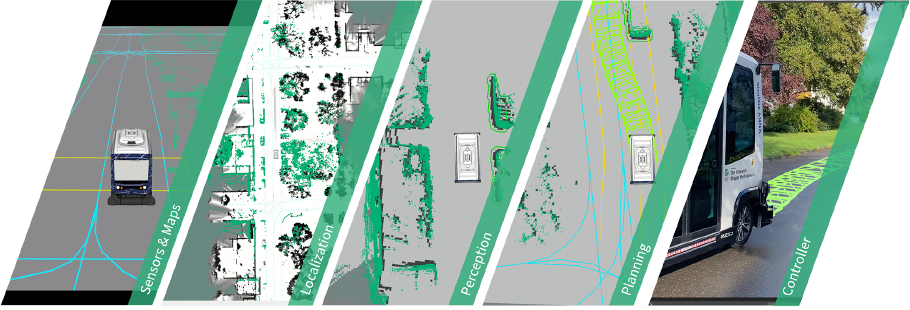}
		\caption{
  The TCS-AD stack provides all the capabilities that are needed for driving on public roads. This ranges from localization, perception, and planning to controller. The performance of the software stack has been demonstrated on various types of vehicles and across multiple test environments.
  \ref{sec:fzi-shuttles} which do not have a steering wheel.
  }
		\label{fig:overview}
	\end{center}
\end{figure}

A multitude of \textit{closed source} driving stacks have been developed in the past. Ta\c{s} et al.~\cite{tas_functional_2016_IV} presented an overview of platforms developed between the early days of autonomous driving up until 2015.
In 2016, the ROS-based \textit{BerthaOne} platform was presented~\cite{tas_bertha_2018_tits}. While the focus was on V2X enabled cooperative driving behaviors within the Grand Cooperative Driving Challenge, the stack is also able to handle urban road traffic. As part of the SAE AutoDrive Challenge, the \textit{Zeus} driving stack was introduced in 2021~\cite{burnettZeusSystemDescription2021}. They demonstrated a variety of defined traffic scenarios, with all algorithms running on CPUs as part of challenge restrictions. The \textit{TiEV} stack competed in the Intelligent
Vehicle Future Challenge 2017, which includes driving on express roads and urban roads~\cite{zhao_tiev_2018}. Similar to \textit{Apollo}, they have developed their software stack from scratch. Marin-Plaza et al. presented a ROS-based stack, which was deployed on two different vehicle types for primarily grid-based off-road environments~\cite{marin-plaza_complete_2018}. Their stack relies extensively on off-the-shelf ROS packages and uses a simple state machine for decision-making.

From the presented driving stacks, patterns emerge. Often, the interchangeability of modules is not considered, which directly hinders collaborative projects. Most industrial stacks are designed with a single vehicle in mind, which introduces scalability issues. Additionally, particular competition requirements are often times at the heart of architectural decisions, which hinders deployment in real-world scenarios. Finally, maturity is also key, enabling the usage by nonexpert people, which expands the testing possibilities in real-world traffic.

 During the development of the software stack, we placed a particular emphasis on well-defined interfaces. These interfaces have been incorporated, particularly in publicly-funded projects, to ensure smooth collaboration with partners. The ability to replace components within the software stack and observe their impact on different vehicles and real-world driving conditions has enhanced our research capabilities. By swapping out individual components, we can isolate and identify the effect of each component on the overall driving performance. This interchangeability has enabled us to conduct controlled experiments to determine the optimal configuration of the software stack in various driving scenarios and to easily include and evaluate newly developed components. Furthermore, we present a baseline implementation for each component, which constitutes a complete stack and reaches a level of maturity such that expert knowledge is not required to operate the test vehicles.

In Sec.~\ref{sec:architecture}, 
we, the group for Technical Cognitive Systems (TCS) at the Research Center for Information Technology (FZI) in Karlsruhe, Germany present our TCS-AD stack and its architecture. In Sec.~\ref{sec:vehicles} we present our research vehicles, followed by demonstrations in Sec.~\ref{sec:demonstrations_applications}. We conclude our work in Sec.~\ref{sec:conclusion}, where we present open research questions.

%% file: sec/3_architecture.tex
\section{ARCHITECTURE}
\label{sec:architecture}

\begin{figure*}[t]
	\centering
        \vspace*{0.2cm}
        \includesvg[width=\textwidth]{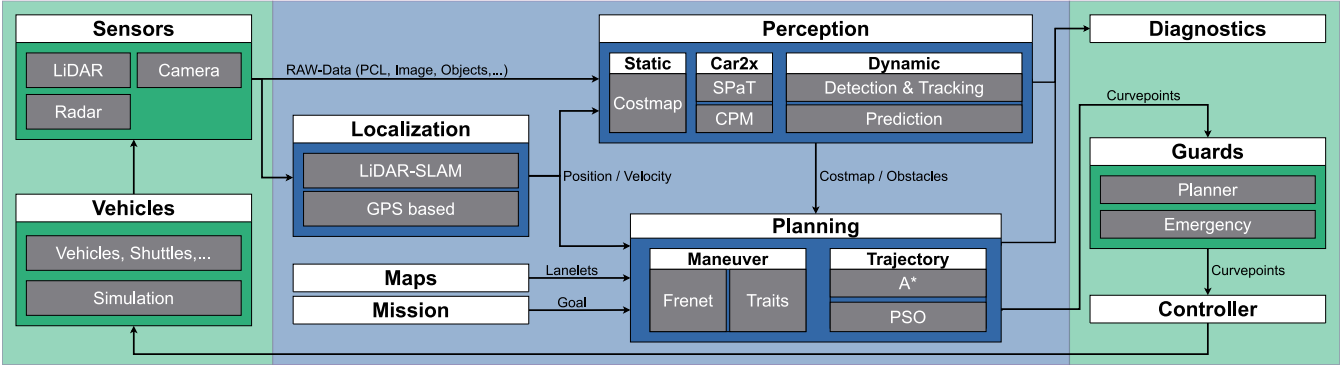}
		\caption{Architecture overview of the TCS-AD function.
        }
		\label{fig:architecture}
\end{figure*}

In this section, the architecture of our automated driving (AD) function and the baseline implementation of each component is presented. 
As visualized in Fig.~\ref{fig:architecture}, the architecture is decomposed into different modules, such as localization, perception and planning. Since we focus on interchangeability and modularity for easy testing of different components, the interfaces and the entire data flow are fixed, while the corresponding components can be easily exchanged.

The localization module is responsible to provide  global position and current velocity information to all other modules where needed, e.g. the perception module and the planning modules. The information is provided in a broadcast fashion as a rigid body transform between the world frame and the vehicle reference frame.

The main design decision in the perception module is the separation of dynamic and static obstacles. Due to the modular system design, different detection and tracking modules have to be used, depending on the sensor setup of the vehicle platform. The different perception algorithms therefore have to deliver a 2D occupancy grid and a list of predicted obstacles. An obstacle is defined with a XYZ position, extends, classification and the predicted positions.

The motion pipeline provides a trajectory in the form of Curvepoints. 
Curvepoints are spatially equidistant sampled trajectory points in Cartesian world coordinates. Additionally, the single points contain a heading angle, a curvature value and a scalar velocity and acceleration. On the top level, the trajectory also contains a timestamp, a strictly monotonically increasing generation number, the number of valid points in the array for support of fixed-size array interfaces and the reference point to convert the Cartesian coordinates into spherical coordinates. Additionally, a car-index adds the possibility to treat the part till the car index either as a prefix or history. This prefix can be used to form higher-order derivatives, which can support the controllers.

All components are reporting their current diagnostic status. The status is one of the following: \textsc{OK}, \textsc{WARN}, \textit{ERROR} and \textsc{STALE}. Additional information can be provided over the message field. This information is processed by the diagnostics module, which is described in Sec.~\ref{sec:diagnose}.

\subsection{High Definition Maps}
\label{sec:maps}

High Definition (HD)-Maps are a crucial part of most environment representations for AD functions. The presented driving stack utilizes the Lanelet format \cite{Bender:2014, poggenhans2018lanelet2}. This representation of the static environment considers geometric as well as topological and semantic aspects.
The key aspects of the Lanelet format are the precise geometric descriptions in geodetic coordinates. These are based on primitives like points, ways, areas and relations. This allows the definition and annotation of semantic information like traffic rules, the definition of relations between lanes and the annotation with arbitrary attributes. 

The map is used by the route planner to determine the optimal route to the destination.
Furthermore, it provides the boundaries of the area, where the vehicle is allowed to drive. This is not necessarily only the actual lane, but can also cover the adjacent lane, if the traffic rules permit a transition.
This is the so-called driving area. It is derived in the form of 2D polygons from the course of the lanes and the related traffic rules.
The left and right side edges limit the driving area in which trajectories may potentially be calculated. This generalization for the trajectory planning allows driving in right-hand as well as left-hand traffic.

\subsection{Localization}
\label{sec:localization}
In order to use the information from the HD-map, a precise global localization of the vehicle is needed. 
The two localization approaches utilized by our software are \emph{Simultaneous Localization and Mapping (SLAM)} and the \emph{Global Navigation Satellite Systems with real-time kinematic (GNSS-RTK)}. Our SLAM algorithm is based on Google cartographer \cite{cartographer} with upcoming additions of utilization of semantic information. Since its publication in 2016, Cartographer has been improved over the last years and the community continues to actively improve and expand the method. Cartographer utilizes a two-stages model: Local scan matching and a global optimization. The global optimizaton searches for the best fit for the local scan matching result 
 
The selection of the actual localization is determined based on the target scenario. For narrow streets or dense environments as encountered in Karlsruhe Weiherfeld-Dammerstock, see upcoming Sec.~\ref{sec:demonstrations_applications}, the SLAM approach is used, but for areas with less dense scenarios GNSS-RTK is utilized.
\subsection{Perception}
\label{sec:perception}

In the following section, we present our environment model used in our driving stack. In CoCar, see Sec.~\ref{sec:cocar}, a commercial LiDAR tracking system consisting out of five IbeoLux LiDAR scanners is available that provides tracked, classified bounding boxes and contour tracks for the {most relevant object classes.
In our FZI-Shuttles, see Sec.~\ref{sec:fzi-shuttles}, a multi LiDAR sensor setup is used with a custom detection and tracking pipeline that consists of the following components: preprocessing, including ground classification (per sensor), pointcloud segmentation (per sensor) and cluster transform, following cluster merging, feature computation, feature tracking and static object removal.

\subsubsection{Preprocessing}
The preprocessing steps for the pointcloud data is utilized by static and dynamic object detection.
\emph{Ground classification} decides per point if the point results from a ground reflection or not. For this, an iterative ground plane estimation technique highly inspired by the paradigm in \cite{zermas2017fast} is used.
\emph{Pointcloud segmentation} clusters points using a range-image based approach inspired by \cite{hasecke2020flic}. As the approach is range-image based and operates on spherical point coordinates instead of Cartesian XYZ coordinates, a clustering per sensor is computed independently.
\emph{Cluster transform} performs a coordinate transformation from sensor coordinate systems to the odometry coordinate frame, which is given by the localization system and the local ego motion estimation.

\subsubsection{Static Obstacles}
\label{subsec:costmap}
The static environment is represented in a 2D occupancy grid. Therefore, an extended version of the base implementation of \cite{costmap_github}, a so called cost-map, is used. The enhancement, for example, comprises the clearing of unknown regions behind obstacles. %
The costmap is built from the ground segmented and transformed pointcloud.

\subsubsection{Detection \& Tracking of Dynamic Objects}
The goal of the Detection \& Tracking components is to provide the locations, extents, classification and kinematic states of relevant objects (pedestrians, bicycles, cars, trucks, etc.) in the surroundings of the vehicle for the prediction, maneuver and motion planning components. Additional to the preprocessing, the following steps are conducted:

\emph{Cluster merging} merges overlapping clusters that relate to point clusters originating from the same objects based on a point cluster distance metric.
\emph{Feature computation} selects features of the object that are used to track the object and estimate the object's kinematics over time. This includes bounding box fitting and L-shape computation based on the closeness-metric based algorithm in~\cite{zhang2017lshape}.
\emph{Feature tracking} associates the computed shape features over time, estimates the vehicle's kinematics using a labeled Gaussian Mixture Probability Hypothesis Density Filter (GMPHD)~\cite{vo2006gaussian,panta2009data} 
\emph{Static object removal} removes all objects from the resulting object list based on a threshold on the estimated velocity of the objects as static objects are already considered in the costmap.

\subsubsection{Prediction}
\label{sec:prediction}

The prediction of the environment is done in a hierarchical order based on \cite{schoerner2021if}.
First, the intentions of the traffic participants are inferred.
The route influences the long term motion of a traffic participant the most.
To obtain estimates about their route, the road users are continuously associated with their corresponding lane using a heuristic consisting of their position and orientation.
The likelihood of the routes is then weighted accordingly to the confidence of the lane matching.
The possible routes are determined considering turn-options up to a certain look-ahead.
Bicycles are assumed to follow along the side of the road whereas vehicles tend to stay in the center.
The longitudinal motion is determined using the Intelligent Driver Model~\cite{Kesting2010}.
The lateral controls are calculated using a Stanley Controller.
These control inputs are then used in an Extended Kalman Filter to predict the next state of the traffic participant. 

Traffic participants that do not follow a certain route are predicted using constant velocity or constant acceleration assumptions.
This applies for example to pedestrians or objects of unknown classification.

An orthogonal approach is used by a graph neural network (GNN), see~\cite{holigraph}.
There, the environment is modelled as a heterogeneous graph which consists of the two node types, namely an agent node and map node.
An agent-node represents a state at a certain time of a traffic participant, e.g., position and velocity.
A map-node represents a segment of a centerline of the HD map.
Context between different nodes is modelled with edges of a certain type.
Edges are also used to model the historic context of a tracked traffic participants, social interaction with others and the connection to HD-map.
The model is able to predict multiple scored trajectories for all traffic participants at the same time.

\subsection{Mission}
\label{sec:mission}
To guarantee the acceptance of only compliant missions into the motion pipeline, a mission control is employed. The mission control checks the incoming requests to avoid situations that are infeasible or could cause the motion pipeline to fail. Most importantly, it guarantees a continuity in local routing. To maximize flexibility, the mission control supports both micro-missions with waypoints and the possibility of appending goals to running missions on the fly. To the outside, the mission control provides an interface, which can then be used with various mission providers, such as a backend of a booking system. This has been shown with a proprietary MQTT-based backend connection as well as the booking backend system. A behavior tree is used to coordinate all preparation steps and the invocation of the motion pipeline.
Additional features that are required for specific use cases can be added to the mission as needed, such as the starting and stopping of a trip recorder that persists relevant sensor data during the autonomous driving mode.

\subsection{Maneuver}
\label{sec:maneuver}

To safely and effectively navigate in complex situations involving dynamic obstacles, we decompose the situation into simpler decision-making units, which we denote as Traits. 
For example, a trait may involve a decision to make the driving area narrower due to oncoming traffic. Another trait wants to extend the driving area due to a parked vehicle on our lane. To find a solution for these contradicting statements, a resolver is implemented. The visualization in Fig.~\ref{fig:trait} shows the algorithm used for resolving the problem. Each trait is represented by a maximum and minimum value, represented as an interval. The interval is also associated with a weight, which is determined trough expert knowledge. The first step of the resolver intersects pairwise all intervals and creates thereby smaller disjunct intervals. We then sum up all weights in the disjunct interval. The final consent is found by linear search in the disjunct intervals for the sign change. The change of sign represents the equilibrium. The result of this example is the extension of the driving area by \SI{2}{\meter}.

\begin{figure}[t]
	\centering
        \vspace*{0.3cm}
	\begin{center}
        \includesvg[width=\columnwidth]{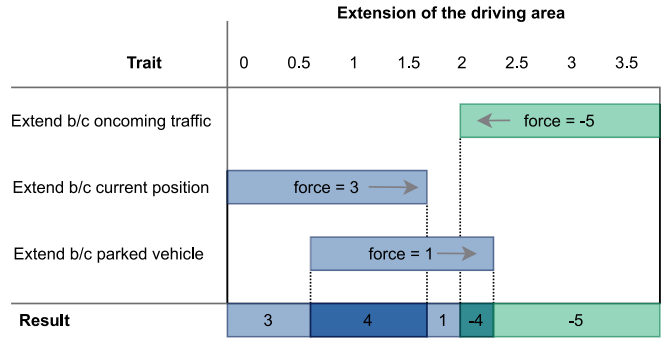}
		\caption{Exemplary visualization for the resolving process for computing the maneuver. In the first step, all intervals are intersected, and the weights are added. The second step includes the search for the equilibrium of the forces. The green interval sets a limit on the maximum extension of the driving area, while the remaining weights vote for an extension, represented in the blue intervals. The consent is found by searching for the change of sign.}
		\label{fig:trait}
	\end{center}
\end{figure}

\subsection{Planning}
\label{sec:planning}
The goal of path planning is to create a safe, collision free trajectory that maximizes passenger comfort. To achieve this goal, we rely on Particle-Swarm-Optimization (PSO). PSO has been shown to be a superior optimization technique compared to other planning algorithms, such as tree-based planning algorithms \cite{darpa2007} that require post-processing smoothing. Additionally, PSO offers advantages over Sequential Quadratic Programming (SQP) \cite{bertha}, which requires a continuously differentiable and convex problem formulation for effective optimization. 

PSO is able to handle arbitrary optimization topologies without any gradient at all. Consequently, this enables the planning algorithm to be used with arbitrary constraints and cost terms, further enhancing the flexibility of the planning process. The trajectory produced by the Particle Swarm Optimization (PSO) algorithm can be described as an equitemporal sequence of poses. The PSO generates these poses by utilizing two constraint concepts: internal and external. Internal constraints refer to the properties that govern the transition between two different poses, such as velocity and curvature. These internal values provide the foundation upon which external constraints are derived. However, due to the inherent complexity of these formulations, they may not always be invertible, which is resolved through the use of discrete differentiation. With these constraints, intelligent sampling strategies are implemented to make the sampling of valid particles feasible. The initial solution is provided by a tree-based pre-planner to find valid paths. The flexibility of this planning architecture has been shown in various applications, such as curvature-rate-limited vehicles as well as multi-axis steering platforms.

\subsection{Execution}
\label{sec:controller}
\subsubsection{Guards}  
Since the nature of the planning process is an optimization problem, there is the possibility that the trajectory violates certain constraints. These constraints are continuously checked by our different guards, which are based on the same architectural structure. 
As input, there is the Curvepoints from the planning process and the current environment from the latest available timestamp.By default, the guards are in the triggered state. The triggered state is initially entered and when a violation of the specific constraints is detected. During this time,  the desired velocity of the incoming Curvepoints are replaced by a braking profile. When the violation is no longer detected, the guard transitions into a state, that applies an acceleration profile from the current velocity of the vehicle to the desired velocity of the planner. 

\subsubsection{Controller}
\label{sec:controler}
The trajectory is then executed by a trajectory controller. 
Given the global position estimate by the localization, the controller projects the vehicle onto the trajectory. 
Depending on the vehicle architecture, the vehicle reference point is either the center of the rear axle, for default Ackermann vehicles, or the center between both axles, for dual steering vehicles. 
From that projection, it calculates the lateral error $e_{lat}$ and angular error $e_{phi}$. Moreover, the curvature, speed and acceleration values are interpolated using temporal look-aheads to compensate for the delay in the actuators.  
The curvature $\kappa$ is used for a feed-forward control using inverse kinematics of a simple bicycle model with the wheelbase $l$. For vehicles with steering rate limits, such as the FZI-Shuttles, see Sec.~\ref{sec:fzi-shuttles}, the controller also monitors the steering velocity. To avoid exceeding the limits, the controller proactively reduces the velocity for sections with high steering velocities, taking the deceleration limits into account. 

\subsection{Diagnostics}
\label{sec:diagnose}

\begin{figure}[t]
	\centering
        \vspace*{0.1cm} 
		\includegraphics[width=\columnwidth]{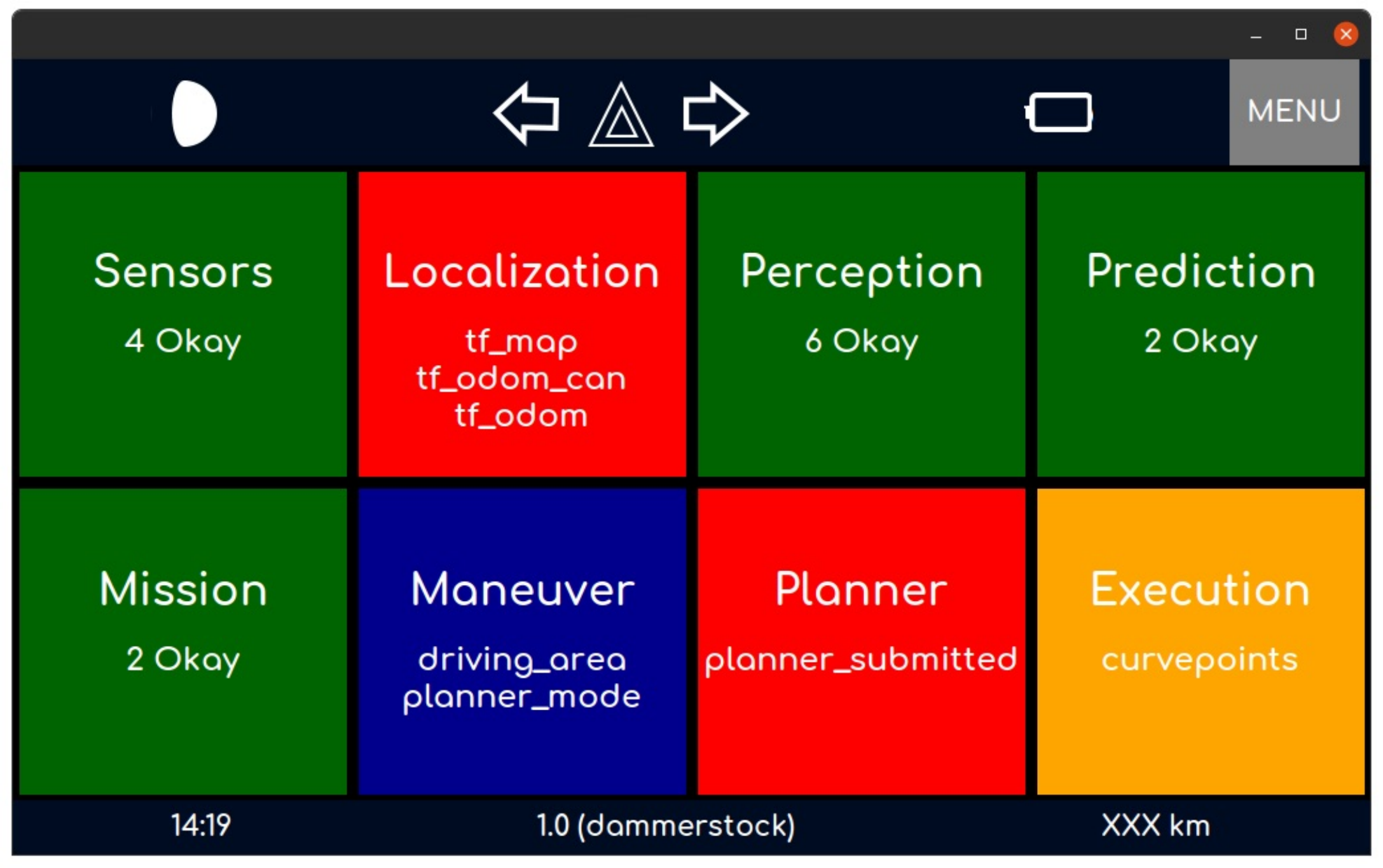}
		\caption{User interface of the diagnostic module. It enables fast debugging of the current state of the AD function. This is especially  useful for the safety operator, which do not have deep insight into the driving function.
        }
		\label{fig:slam-map}
\end{figure}

In complex systems, such as autonomous vehicles, the ability to identify and track failures during operation is of utmost importance. 
To accomplish this, various aspects of the system are monitored, typically through observation of the components' behavior or the use of self-states. These monitoring methods provide insights into the inner workings of the system and allow operators to pinpoint any anomalies that may arise. One way to observe the system is to measure the absence of data, as well as their frequency and delay. The data and corresponding meta-data are then analyzed to detect any abnormal behavior. The detection can be data based or model based driven. The used models can be constructed using either manual techniques or machine learning approaches. For example, the localization component is diagnosed by statistically modelling its behavior based on odometry \cite{OrfMFILocalizationUncertainty2023}.  

It is inevitable that failures occur during operation for example due to hardware degradation. In such cases, it is crucial to provide timely and accurate information to operators about the nature of the anomaly, using abstracted error descriptions that allow for appropriate countermeasures to be taken. The TCS-AD stack is suitable to both applications by recognizing and addressing failures on a technical level, and then aggregating the data to provide a summary that is abstracted and easier to comprehend for operators. 

\subsection{Vehicle2X}
\label{sec:v2x}

Vehicle2X (V2X) communication enables real-time data sharing, cooperative driving and promises to increase safety. MAP Data Messages (MAP) and Signal Phase and Timing Messages (SPaT) are used to receive intersection and traffic light information. This information is used in combination with HD maps for anticipatory driving and braking. 
Sensed obstacles and traffic participants are broadcast between the AD stack and smart infrastructure both ways via Collective Perception Messages (CPM) to enrich the environment model of both systems. Automated vehicles can utilize these as extensions of the vehicle's sensors for safer driving, especially in intersections with limited visibility. These use-cases are already presented with the support of the Test Area Autonomous Driving Baden-W\"urttemberg (\mbox{TAF-BW}).
V2X also supports to achieve automation levels that do not require safety drivers through online monitoring and controlling of the vehicle from remote. I.e. the national legal regulation in Germany \cite{bmdv} requires a technical supervision which can also be remote.

%% file: sec/4_vehicles.tex
\section{TESTING VEHICLES}
\label{sec:vehicles}

\begin{figure*}[t]
        \vspace*{0.3cm}
	\begin{center}
        \begin{subfigure}[T]{0.5\columnwidth}
            \centering
			\includegraphics[width=\columnwidth]{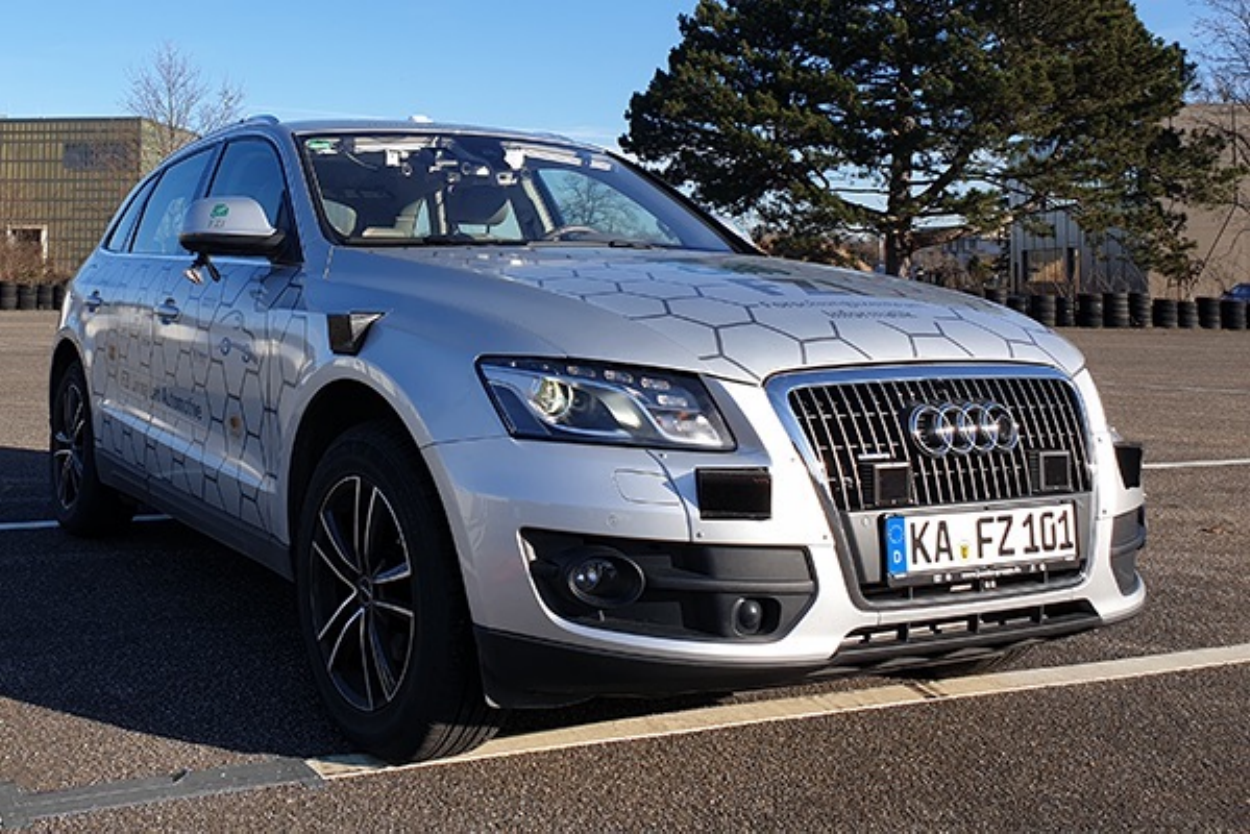}
			\caption{CoCar is based on an Audi Q5 and was build up 2011.}
			\label{fig:cocar}
		\end{subfigure}
        \begin{subfigure}[T]{0.5\columnwidth}
            \centering
    		\includegraphics[width=\columnwidth]{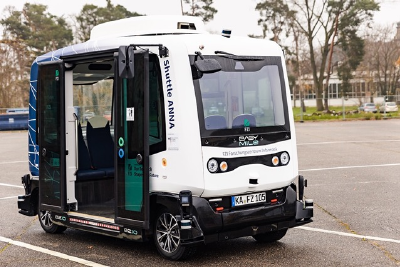}
    		\caption{FZI-Shuttles are based on EasyMile-EZ10 Gen2.
      and used for on-demand passenger transport since 2019.
      }
    		\label{fig:fzi-shuttle}
		\end{subfigure}
		\begin{subfigure}[T]{0.5\columnwidth}
            \centering
    		\includegraphics[width=\columnwidth]{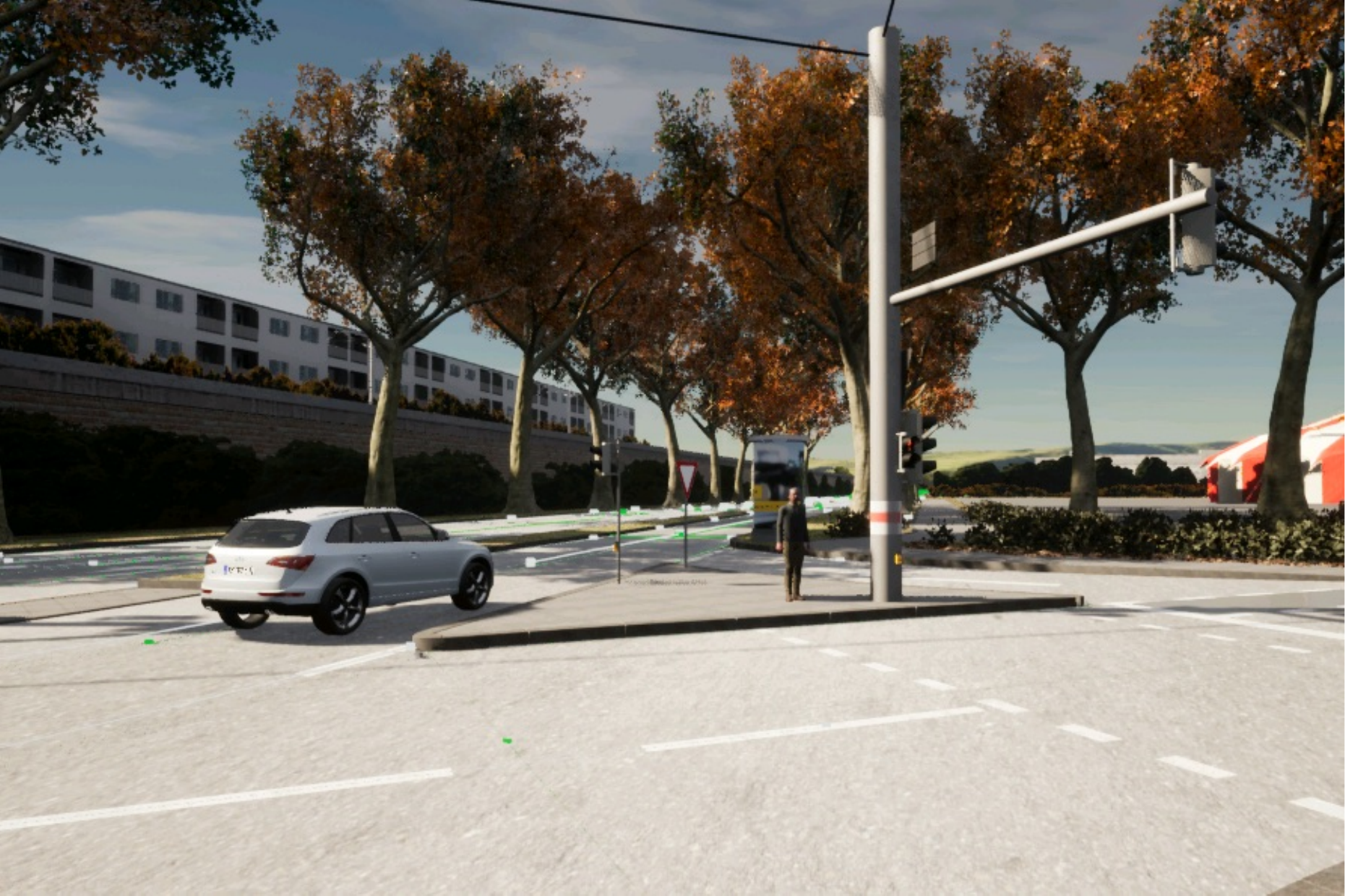}
    		\caption{Simulator CARLA with CoCar and pedestrians on an intersection in Karlsruhe.}
    		\label{fig:simulation}
        \end{subfigure}
		\label{fig:vehicles}
	\end{center}
 \caption{All test vehicles  of the FZI are presented. They were all equipped at the time of construction with state-of-the-art LiDAR system and D-GPS sensors.}
\end{figure*}

The TCS-AD Stack was deployed in several public funded projects.
Especially on our own FZI vehicles, the stack is constantly tested and improved. In the following, we want to give a brief overview over the vehicles.

\subsection{CoCar}
\label{sec:cocar}

Our testing vehicle CoCar (Cognitive Car) is based on an Audi Q5 and is equipped with LiDAR, GPS and Camera systems. The sensor setup consists of a high-precision GNSS/INS provided by Oxford Technical Solutions (OxTS RT3000), five Ibeo Lux sensors with 4 Layers to provide a 360 degree surround view. The road approval of CoCar allows us to test our autonomous driving function in all areas in Germany.

\subsection{FZI-Shuttles}
\label{sec:fzi-shuttles}

The FZI-Shuttles ELLA and ANNA are based on EasyMile EZ10 Gen2 platforms. 
The shuttles are able to steer with both axes and therefore have a high lateral mobility. Since the driving stack is not bound to virtual rails but to move freely in everyday traffic, the platforms were modified to fit our needs. This includes the mounting of six LiDAR sensors. %

\subsection{Simulation}

In addition to real road testing, we use simulations to develop and validate our driving stack and for testing individual components. 
The simulation data such as environmental information, sensor data and positional data of other road users is used as input for the AD-Stack and processed as on the real vehicle to plan the trajectory and control the vehicle.
For validation of interactions with other road users, we use different methods. For instance, we use SUMO \cite{sumo} in co-simulation with CARLA \cite{carla} to simulate high traffic volume. We also define scenarios in the ASAM OpenSCENARIO format\cite{ASAM_SIM_GUIDE} to achieve very specific behavior of other road users. Due to the repetitive behavior, we are able to analyze, fine-tune and evaluate our AD stack. Additionally, using digital twins of out testing areas and vehicles, as shown in Fig.~\ref{fig:simulation}, we are able to early integrate and test the AD stack before going into the site itself. This also enables endurance tests to be carried out. Road users such as pedestrians can be immersed in the simulation using virtual reality. In this way, it is possible to familiarize people with the AD-Stack and to research interactions in a very realistic way.

%% file: sec/5_demo.tex
\section{DEMONSTRATIONS AND APPLICATIONS}
\label{sec:demonstrations_applications}

\begin{figure}[ht]
	\begin{center}
        \begin{subfigure}[T]{0.49\columnwidth}
            \centering
			\includegraphics[width=\columnwidth]{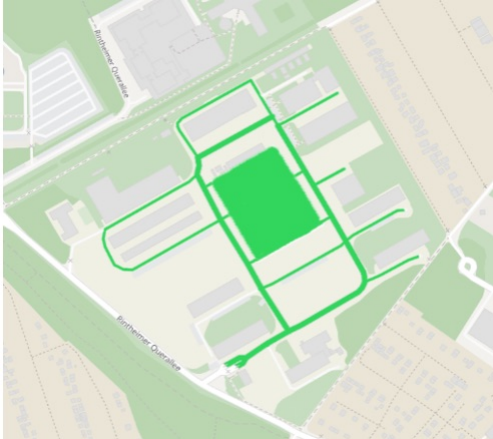}
 		  \caption{The KIT Campus East is a restricted area with about \SI{2}{\kilo\meter} of road and \SI{5000}{\meter\squared} free space.}
		  \label{fig:campus_ost}
		\end{subfigure}
        \begin{subfigure}[T]{0.49\columnwidth}
            \centering
		  \includegraphics[width=\columnwidth]{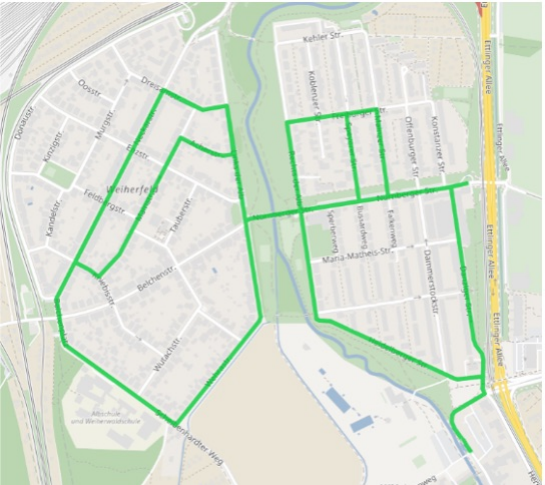}
		  \caption{Weiherfeld-Dammerstock with a street network of about \SI{6.4}{\kilo\meter}}
		  \label{fig:wd}
		\end{subfigure}
  \caption{Both main test sites, KIT Campus East and Weiherfeld-Dammerstock (WD), are presented. The KIT Campus East is used for the integration of automated driving functions and initial tests. Afterward, the tests are conducted in WD, where we are interacting with mixed traffic-flows and VRUs. Background for images provided by \cite{OpenStreetMap}.}
		\label{fig:test_sites}
	\end{center}
\end{figure}

The automated driving stack is mainly evaluated in and around Karlsruhe, Germany in urban and suburban areas. Two selected locations are presented in the following. The first one is a restricted area that is part of the Karlsruhe Institute of Technology (KIT), the so-called Campus East. It is part of the TAF-BW, dedicated to early rapid prototyping vehicle tests. The 2nd test site is Weiherfeld-Dammerstock, an urban area with various types of public roads. The challenges faced there are narrow streets, parked vehicles along the road, intersections with poor visibility, and interaction with VRUs. We autonomously drove over \SI{3000}{\kilo\meter} with our testing vehicles in these areas. Also, the FZI-Shuttles were employed in regular passenger transport, operated by safety drivers from the local public service provider. This required a very high quality and safety standard, as these safety operators do not have the same insights as developers have and passenger transport is only possible when achieving high comfort and safety standards.
Also, the stack and its modules are used in multiple projects with different partners. Because of this reason, the stack also shows its adaptability and flexibility. It follows a small excerpt of the projects: RobustSense \cite{robustsense-hp}, SmartLoad \cite{smartload-hp}, EVA-Shuttle \cite{EVA-shuttle-hp}, SafeADArchitect \cite{safead-architect}, VVM \cite{vvm-hp}, SmartEPark \cite{smartepark-hp}, Shuttle2X \cite{s2x-hp}, \"OV-Leitmotiv-KI \cite{oev-leitmotif-hp}, KIsSME \cite{kissme-hp}, KIGLIS \cite{kiglis-hp}.

%% file: sec/6_conclusion.tex
\section{CONCLUSION}
\label{sec:conclusion}

We presented a modular architecture for automated driving in this paper, which can be adapted to various vehicles and simulation environments. With this modular design, we can deploy new components on the vehicles easily and evaluate them in restricted and public areas with mixed traffic. Our test vehicles drove autonomously for over \SI{3000}{\kilo\meter} in these environments. We are able to incorporate machine learning-based algorithms in various parts of the software stack, because our model based trajectory planner outputs only drive-able and compliant trajectories.  Improvements in our algorithms will also enable us to advance in more complex traffic scenarios. We aim to develop algorithms that can eliminate the need for safety operators at all, or at least from the inside of the vehicle. This will involve designing and testing advanced autonomous systems that can ensure safe and reliable operations without human intervention.